\documentclass[10pt]{article}

\usepackage[a4paper,margin=1in]{geometry}
\usepackage[T1]{fontenc}
\usepackage[utf8]{inputenc}
\usepackage{lmodern}
\usepackage{mathptmx}
\usepackage{microtype}
\usepackage{amsmath,amssymb,amsthm}
\usepackage{graphicx}
\graphicspath{{figures/}}
\usepackage{booktabs}
\usepackage{multirow}
\usepackage{array}
\usepackage{xcolor}
\usepackage{enumitem}
\usepackage[round,authoryear]{natbib}
\usepackage[colorlinks=true,allcolors=blue!60!black]{hyperref}

\title{A Decomposable Probe for Few-Step Diffusion Models:\\
Prompt, Latent, and Score Selectivity across Backbone Families and Distillation Paradigms}
\author{Patrick Mu Haojie\thanks{Independent Researcher. \texttt{patrickmupersonal@outlook.com}}}
\date{}

\begin{document}

\maketitle

\begin{abstract}
Few-step distilled diffusion students cut text-to-image inference from
$\sim\!50$ to $1$--$8$ network evaluations, but the resulting quality gap
is almost always summarised by a single end-to-end FID/CLIP scalar that
cannot say \emph{which} axis of the conditioning response changed, nor
whether a behaviour is a property of the architecture, the distillation
objective, or simply of being a diffusion model. We replace the scalar
with a decomposable probe that injects controlled perturbations
along three layers (prompt encoder, denoiser input, denoiser output)
under three modes (mean, variance, scale) and six strengths, and reports
a bootstrap-median Bures $W_2^2$ selectivity ratio on Inception
features. Applied under a single matched estimator to a sweep of $23$
models --- five teachers and $18$ distilled students across five
backbone families (SDXL, SD1.5, SD3.5, PixArt-$\alpha$, FLUX), three
architecture classes (UNet, DiT, MMDiT), and five distillation
paradigms --- the three layers read three empirically separable factors: the prompt
layer is a universal prompt-mean response within this diffusion T2I sweep (a sanity channel rather than a discriminator), the latent
layer reads the prediction type, and the score layer reads the
distillation objective. Our main result is that, within this sweep, the latent layer is a
\emph{near-binary detector of rectified-flow
backbones}: its ratio exceeds $1$ across a sustained low-to-mid band only
for rectified-flow models (SD3.5, FLUX) and no
$\epsilon$-prediction model satisfies the sustained-band criterion; the
obvious wide-T5-conditioning explanation is ruled out
by a matched $\epsilon$-prediction control (PixArt-$\alpha$), and the
fingerprint survives adversarial (ADD) distillation on two RF
backbone families as both teacher and student. Two secondary score-layer
findings hold across families under narrower scopes: a canonical
$4$-step Turbo-family ADD-vs-rest contrast (the ADD student is the lowest
same-family student at the $4$-step comparison on the two UNet families
that have a non-ADD baseline --- not extrapolated to $1$-step, where
aggressive non-ADD students can enter the ADD range; the two
rectified-flow families contribute ADD-only points consistent
with the low-ADD end but without a same-family non-ADD contrast), and a CI-separated trajectory-rollout early-strength score
spike observed on
both UNet and DiT, with magnitude depending on step count and backbone.
All ratios are CI-citable under one estimator; we
release the per-cell tables and the estimator.
\end{abstract}

\section{Introduction}
\label{sec:intro}

Few-step distilled diffusion students --- ADD/Turbo, Lightning, LCM,
DMD2, Hyper-SD, and their kin --- have collapsed text-to-image inference
from $\sim\!50$ network function evaluations down to $1$--$8$. The price
of this speedup is well documented in aggregate (worse FID, weaker
prompt alignment, reduced sample diversity), but is reported almost
exclusively as a single end-to-end scalar that treats the student as a
black box. Such a scalar cannot distinguish a student that has lost
prompt-mean sensitivity from one whose latent or score response has
changed, nor can it say whether a given behaviour is a property of the
\emph{architecture}, the \emph{distillation objective}, or simply of
being a diffusion model at all. These are mechanistically distinct
properties, and a headline number cannot tell them apart.

We take the opposite approach: instead of one scalar, a
\textbf{decomposable probe}. Our diagnostic injects controlled
perturbations into the forward pass along three orthogonal
\emph{layers} --- the prompt encoder, the denoiser input (latent), and
the denoiser output (score) --- under three \emph{modes} (mean,
variance, scale) and six strengths, and summarises each cell of the
resulting grid by a bootstrap-median Bures $W_2^2$ selectivity ratio on
Inception features. The probe is training-free, acts through forward
hooks, and is backbone-agnostic.

\paragraph{The three layers read three empirically separable factors.}
Applied under a single matched estimator to a sweep of $23$ models ---
five teachers and $18$ distilled students, spanning five backbone
families (SDXL, SD1.5, SD3.5, PixArt-$\alpha$, FLUX), three architecture
classes (UNet, DiT, MMDiT), and five distillation paradigms --- the
three layers turn out to be far from redundant. Each reads a distinct,
empirically separable property of the model (we use ``orthogonal''
descriptively throughout, not as a claim of statistical independence):
the \textbf{prompt} layer is a
\emph{universal} prompt-mean response within this diffusion T2I sweep (every model is
prompt-mean selective and so it discriminates none of our factors of
interest, making it a
sanity channel rather than a discriminator); the
\textbf{latent} layer is a near-binary detector of the \emph{prediction
type} (rectified-flow vs $\epsilon$-prediction); and the \textbf{score}
layer reads the \emph{distillation objective}. This decomposition is the
spine of the paper and the lens through which we read three findings.

\paragraph{Finding 1 (main): the latent layer detects rectified-flow,
and the fingerprint survives ADD distillation.} The latent layer shows a sustained
low-to-mid-strength elevation ($R$ above $1$ across a band of strengths)
only for the two rectified-flow backbones (SD3.5, FLUX);
$\epsilon$-prediction models either stay below $1$ or show only isolated,
non-band-shaped excursions. A control disambiguates the
cause: PixArt-$\alpha$ shares SD3.5's wide T5 conditioning but is
$\epsilon$-prediction, and it does not reproduce the band, ruling out
the obvious wide-T5-conditioning explanation and attributing the effect to the
rectified-flow denoiser. The fingerprint moreover survives distillation:
the two distilled students on RF backbones --- both pure-adversarial
(ADD), an objective that flattens the score-layer signal --- still pass
on the latent layer. Within this sweep the latent layer is thus a
near-binary detector of the prediction type (read as a sustained
low-to-mid band) that survives ADD distillation, seen across two
RF backbone families (MMDiT and DiT), as both teacher and student.

\paragraph{Finding 2: a score-layer ADD-vs-rest detector.} On the score
layer at the canonical strength $s\!=\!0.5$, and \emph{at the canonical
$4$-step comparison}, the adversarial-distilled (ADD) student is the lowest of its
family, below every $4$-step non-ADD student with non-overlapping
confidence intervals on both UNet families that contain one; the two
non-UNet ADD students extend the low-ADD end to MMDiT and DiT. We state
this as a binary contrast at the $4$-step comparison and explicitly \emph{not} as a
five-paradigm ranking: the four non-ADD paradigms are not pairwise
separable at this cell, and this should not be extrapolated to $1$ step,
where the most aggressive non-ADD
students can enter or even undercut the ADD range (SDXL Hyper-1 $0.35$
below Turbo-1 $0.49$).

\paragraph{Finding 3: a trajectory-rollout early-strength score spike.}
Students whose loss includes a multi-step trajectory-rollout regression
term, without a dominant adversarial component, exhibit a localized
early-strength spike on the score layer: the low-strength peak
($s\!\leq\!0.1$) has a confidence interval strictly above $1$ and
CI-separated above the value at $s\!=\!0.5$. This appears for consistency
(LCM) and backward-simulation (Flash) students on UNet and most strongly for LCM
on a pure DiT backbone, while the DiT/MMDiT ADD controls (SD3.5-Turbo,
FLUX-schnell) show no such
spike. The spike is associated with trajectory-rollout objectives and
observed on both UNet and DiT, but its appearance and magnitude depend
on step count and backbone.

\paragraph{Scope.} This paper is an instrument and its readings. We
report the probe and what it reveals across the sweep; we do
\emph{not} attach a downstream task (e.g.\ model attribution or a
training-time correction) here, and we are deliberate about the limits
of each reading --- the score-layer finding is a binary detector, not a
paradigm ranking. We release all per-cell CI-citable tables and the
matched estimator.

\paragraph{Contributions.}
\begin{itemize}[leftmargin=1.2em,topsep=2pt,itemsep=2pt]
  \item \textbf{A decomposable layer-/mode-resolved probe} for
        diffusion text-to-image models, with a bootstrap-median true
        Bures $W_2^2$ selectivity ratio and a single matched estimator
        applied identically to $23$ models.
  \item \textbf{A separable-factor decomposition}: across five
        families and three architecture classes, the prompt / latent /
        score layers read a universal property / the prediction type /
        the distillation objective, respectively.
  \item \textbf{A rectified-flow detector}: within this sweep the latent layer is a
        near-binary detector of the prediction
        type (read as a sustained low-to-mid band) that survives ADD distillation,
        with the obvious wide-T5-conditioning explanation ruled out by a matched
        $\epsilon$-prediction control and four RF cases spanning two
        architecture families and teacher/student roles.
  \item \textbf{Two cross-family score-layer findings}: a canonical
        $4$-step Turbo-family ADD-vs-rest contrast (ADD lowest same-family
        student at the $4$-step comparison on the two UNet
        families, not extrapolated to $1$-step, with ADD-only points extending it to the two RF
        families), and a trajectory-rollout
        early-strength spike present on both UNet and DiT.
\end{itemize}

\section{Related Work}
\label{sec:related}

\subsection{Few-Step Diffusion Distillation}
\label{sec:related:distill}

Recent years have produced a dense menu of methods that compress
multi-step diffusion teachers ($20$--$50$ sampler steps)
\citep{ho2020ddpm,song2021ddim} into
1--8-step students. We group them by
the distillation \emph{loss family}, since this taxonomy is the
backbone of our experimental design rather than a survey end-point:
\textbf{(a) Adversarial Diffusion Distillation} (ADD)
\citep{sauer2023add} and its progressive variant SDXL-Lightning
\citep{lin2024lightning} train the student against an
adversarial discriminator and a teacher-derived score loss;
\textbf{(b) Consistency Models} \citep{song2023consistency,
song2023improvedcm} and their latent-space adaptations LCM and
LCM-LoRA \citep{luo2023lcm,luo2023lcmlora} train the student to be
self-consistent along the diffusion trajectory;
\textbf{(c) Score Identity / Distribution Matching Distillation}
(SiD, DMD, DMD2) \citep{zhou2024sid,yin2023dmd,yin2024dmd2}
reformulate the distillation objective as an identity over scores or
as a distribution-matching loss with an auxiliary fake-score network;
\textbf{(d) Trajectory-segmented hybrids} such as Hyper-SD
\citep{ren2024hypersd} interleave consistency, score, and adversarial
objectives on partitioned timestep ranges; and
\textbf{(e) Backward simulation} approaches like Imagine Flash
\citep{kohler2024imagineflash} and Flash Diffusion
\citep{chadebec2024flash} feed student trajectories back into
the training loop. Our sweep instantiates these families on non-UNet
backbones as well --- SD3.5-Turbo (ADD on MMDiT,
\citealp{esser2024sd3,sd35turbo2024card}),
FLUX.1-schnell (ADD on DiT, \citealp{flux2024,fluxschnell2024card}), and PixArt-LCM
(consistency on DiT, \citealp{chen2024pixart,chen2024pixartdelta,luo2023lcm}) --- which is
what lets the probe's readings be tested off the UNet backbone.

Table~\ref{tab:zoo} uses short operational paradigm labels that map onto
these families: \emph{ADD}~(a) covers both the Turbo-family adversarial
objective (SD-Turbo, SDXL-Turbo) and the progressive-adversarial variant
SDXL-Lightning \citep{lin2024lightning} --- which builds on the
progressive-distillation lineage \citep{salimans2022progressive} but adds
an adversarial objective --- and which we label \emph{progressive-adv} to
distinguish it from the Turbo-family ADD variant;
\emph{trajectory} denotes the consistency family~(b) (LCM);
\emph{distribution-match}~(c); \emph{mixed} denotes the
trajectory-segmented hybrid Hyper-SD~(d); and Flash Diffusion~(e) is a
backward-simulation objective, labelled \emph{backward-sim}. We use the
term \emph{trajectory-rollout} operationally in
Section~\ref{sec:exp:spike} for objectives that train on student or
generated trajectory segments --- the consistency~(b) and
backward-simulation~(e) families --- even when combined with auxiliary
adversarial terms, and reserve \emph{ADD} for the adversarial-dominant
objectives~(a).

\subsection{Architectures and Prediction Types}
\label{sec:related:arch}

Our sweep spans three architecture classes that differ along two axes
relevant to the probe: the backbone family and the denoiser prediction
type. \textbf{(i) UNet with $\epsilon$-prediction}: SDXL
\citep{podell2024sdxl} (two CLIP \citep{radford2021clip} text encoders)
and SD1.5 (one CLIP encoder) are the classical latent-diffusion
\citep{rombach2022ldm} configuration. \textbf{(ii) Diffusion Transformer (DiT) with
$\epsilon$-prediction}: PixArt-$\alpha$ \citep{chen2024pixart} replaces
the UNet with a pure transformer \citep{peebles2023dit} and conditions on
a single T5-XXL \citep{raffel2020t5} encoder, holding the wide
T5 conditioning of the rectified-flow models fixed while keeping
$\epsilon$-prediction --- which is exactly what lets it serve as the
control that isolates the prediction type in
Section~\ref{sec:exp:rf-detector}. \textbf{(iii) Rectified-flow
denoisers}: SD3.5 (an MMDiT with two CLIP encoders plus T5,
\citealp{esser2024sd3}) and FLUX \citep{flux2024}
(a DiT with CLIP-L plus T5)\footnote{FLUX.1 uses a hybrid transformer
with both double-stream (MMDiT-style) and single-stream blocks; we write
``DiT'' as shorthand and treat SD3.5 and FLUX as two rectified-flow
backbone families rather than as fully independent architectures.} are
trained under a flow-matching /
rectified-flow objective \citep{liu2023rectflow, lipman2023flowmatching}
rather than $\epsilon$-prediction. The
prediction-type axis (rectified-flow vs $\epsilon$-prediction) is the
one the latent layer turns out to read.

\subsection{CFG Geometry in Multistep Diffusion}
\label{sec:related:cfg}

A separate thread analyses the geometry of classifier-free guidance
(CFG, \citealp{ho2022classifierfree}) and its effect on sample diversity.
\citet{bradley2024classifier} show that CFG behaves as a
predictor--corrector on the score, alternating a denoising predictor
with a sharpening corrector in the SDE limit, and
\citet{karras2024guidance} disentangle quality from variation by
guiding generation with a weaker version of the same model
(autoguidance) rather than with an unconditional one. These analyses
are multi-step and live in the limit of large NFE. Our setting differs
in two respects: we operate on \emph{distilled few-step} students
(1--8 NFE) and we measure response geometry on real Inception features
rather than in an analytical large-NFE limit.

\subsection{Inference-Time / Test-Time Interventions}
\label{sec:related:inference}

A group of methods modifies inference without altering diffusion
weights. PAG (perturbed-attention guidance,
\citealp{ahn2024pag}) replaces self-attention with identity in a
guidance branch; FreeU \citep{si2024freeu} reweights the U-Net
backbone features while suppressing low-frequency skip-connection
components; AYS (align-your-steps, \citealp{sabour2024ays}) reshapes
the timestep schedule under a KL-optimality criterion. We relate to
this line methodologically: like these methods our probe acts through
forward hooks and needs no retraining, but it is a \emph{diagnostic
instrument} that reads model behaviour rather than an intervention that
changes it.

\subsection{Diagnostic Metrics for Generative Models}
\label{sec:related:metrics}

The metric stack we use is standard but combined deliberately around
one design choice the related literature does not commit to: a
\emph{bootstrap-median Bures $W_2^2$ ratio} as the headline statistic.

The \emph{full} closed-form Gaussian $W_2^2$ on Inception features
is the natural feature-space distance under our setup but is
non-linear in the empirical covariance, so the plug-in estimator
does not coincide with the bootstrap distribution of the same
functional at our cell sizes (cf.\ Section~\ref{sec:method:k1}). We
therefore report a bootstrap-median ratio with the full Bures $W_2^2$
functional inside the bootstrap loop; the bootstrap-median aligns the
point estimate and its confidence interval under a single statistical
object. Larger-scale alternatives such as Sinkhorn--$W_2$
\citep{cuturi2013sinkhorn} could be substituted without changing the
qualitative conclusions. We read single-number FID with care, since it
is a biased estimator whose bias depends on the model and sample size
\citep{chong2020fidinfinity, heusel2017fid}; our headline statistic is
therefore the bootstrap-median ratio rather than a raw plug-in FID. A
$W_2^2$-independent diversity channel (e.g.\ a Vendi score
\citep{friedman2023vendi} on DINOv2 \citep{oquab2024dinov2} embeddings,
which does not share
the Bures functional with our headline ratio) is a natural complement
to the probe; we leave a full diversity readout to future work.

\section{The Probe}
\label{sec:method}

\begin{figure}[t]
\centering
\includegraphics[width=\linewidth]{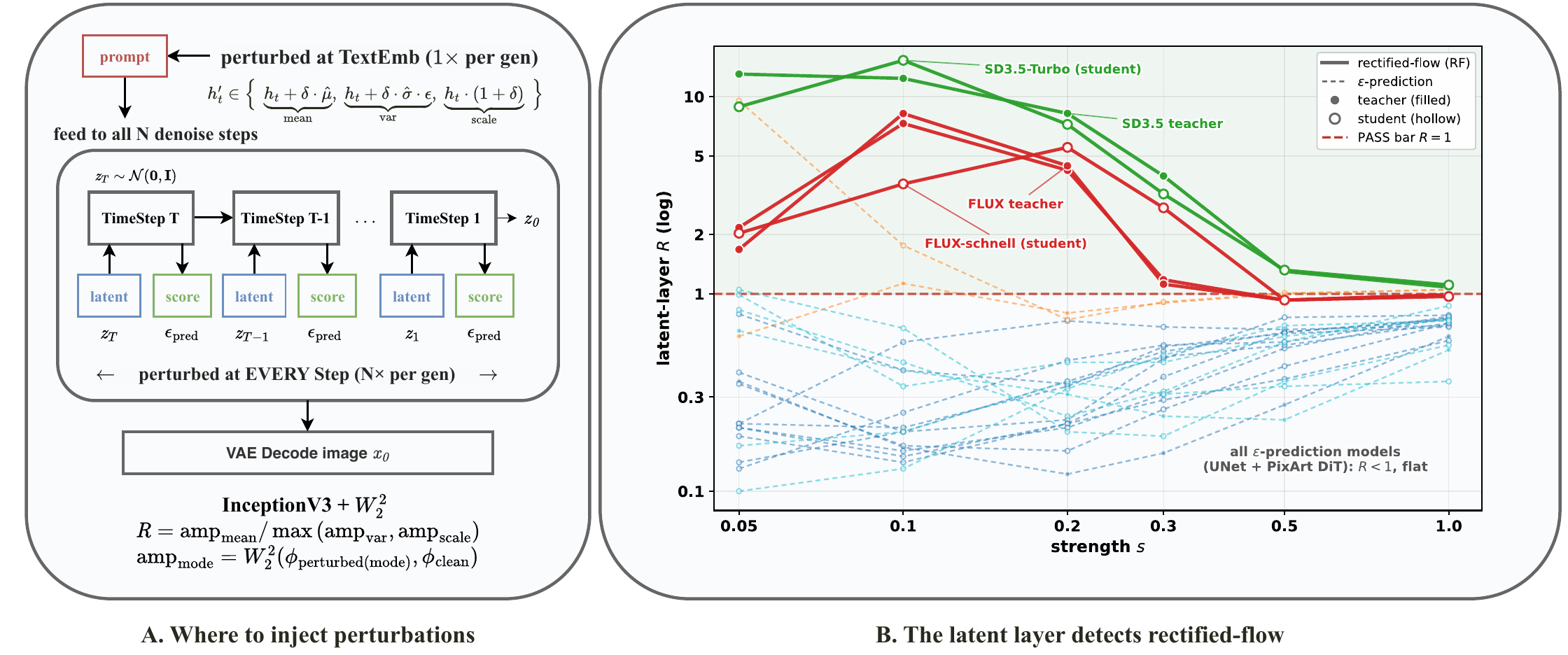}
\caption{\textbf{The probe and its main reading.}
\textbf{(A)} Where and how the probe injects. The prompt embedding is perturbed
once per generation at the text encoder; the latent and score tensors
are perturbed at every one of the $N$ denoise steps. Each layer is hit
under three modes --- mean ($h_t{+}\delta\hat{\mu}$), variance
($h_t{+}\delta\hat{\sigma}\epsilon$), and scale ($h_t(1{+}\delta)$) ---
across six strengths. The decoded image is embedded by Inception-v3 and
each cell is summarised by the matched selectivity ratio
$R = \mathrm{amp}_\text{mean} / \max(\mathrm{amp}_\text{var},
\mathrm{amp}_\text{scale})$, with
$\mathrm{amp}_\text{mode} = W_2^2(\phi_{\text{perturbed(mode)}},
\phi_\text{clean})$.
\textbf{(B)} The main finding (Section~\ref{sec:exp:rf-detector}). On
the latent layer, $R$ stays above the $R{=}1$ equality line (where the
mean perturbation matches the strongest higher-moment perturbation)
across a sustained low-to-mid band \emph{only} for the two rectified-flow
backbones (SD3.5, FLUX), as both
teachers (filled) and distilled students (hollow); $\epsilon$-prediction
models (the UNet families and the PixArt-$\alpha$ DiT) do not form the
band --- they stay below $1$ or show only isolated low-$s$ excursions.
Within this sweep the latent layer thus acts as a near-binary
empirical fingerprint of the prediction type that survives ADD
distillation on the two RF backbones tested.}
\label{fig:probe-rf}
\end{figure}

We probe a diffusion text-to-image model by injecting controlled
perturbations into its forward pass and measuring how the output
distribution responds. The probe is deliberately \emph{decomposable}:
it factors the response along three injection \emph{layers} and three
perturbation \emph{modes}, and (Section~\ref{sec:exp:orthogonal}) each
layer turns out to read a different, orthogonal property of the model.

\subsection{Layer--Mode Selectivity Scan}
\label{sec:method:k1}

\paragraph{Setup.}
Given a diffusion model $G_\theta$ (teacher or distilled student), we
inject controlled perturbations along three orthogonal \emph{layers}
and measure how the output distribution responds. The three layers are:
\begin{itemize}[leftmargin=1.2em,topsep=2pt,itemsep=1pt]
  \item \textbf{Prompt layer} ($\mathcal{P}_\text{prompt}$): forward hook on
        the text-encoder hidden states (one site per encoder; backbones
        expose 1--3 text encoders, see Section~\ref{sec:exp:setup}).
  \item \textbf{Latent layer} ($\mathcal{P}_\text{latent}$): forward
        pre-hook on the denoiser input (UNet for SDXL/SD1.5; the
        DiT/MMDiT \texttt{transformer} for PixArt/SD3.5/FLUX).
  \item \textbf{Score layer} ($\mathcal{P}_\text{score}$): forward hook on
        the denoiser output (the predicted $\epsilon$/velocity).
\end{itemize}
On each layer we apply three perturbation \emph{modes} of strength $s$:
\textbf{mean} ($x \mapsto x + s\hat\mu$), \textbf{variance}
($x \mapsto x + s\hat\sigma\,\varepsilon$, $\varepsilon\sim\mathcal{N}(0,I)$),
and \textbf{scale} ($x \mapsto (1+s)x$),
where $\hat\mu,\hat\sigma$ are estimated on a calibration set.
The full grid is $3\,\text{layers}\times 3\,\text{modes}\times 6\,\text{strengths}$
with $s\in\{0.05,0.1,0.2,0.3,0.5,1.0\}$.

\paragraph{Selectivity ratio.}
For each (layer, strength) cell, let
$\mathrm{amp}_m \;=\; W_2^2\!\bigl(\phi(G_\theta(\cdot;\,\mathrm{perturb}_m)),\,\phi(G_\theta(\cdot))\bigr)$
denote the squared $W_2$ distance between perturbed and clean Inception-v3
\citep{szegedy2016inception} features $\phi$, evaluated under perturbation
mode $m\in\{\text{mean},\text{var},\text{scale}\}$.
Each $\mathrm{amp}_m$ is computed using the full Bures $W_2^2$
functional \citep{bhatia2019bures} on $d\!=\!2048$ Inception-v3 features,
evaluated within a
bootstrap-median framework (Section~\ref{sec:related:metrics}).
We summarise the cell by the \emph{selectivity ratio}
\begin{equation}
\label{eq:ratio}
  R_\text{sel} \;=\; \frac{\mathrm{amp}_\text{mean}}{\max(\mathrm{amp}_\text{var},\,\mathrm{amp}_\text{scale})},
\end{equation}
which is large iff the mean perturbation dominates both higher-moment
perturbations on that layer.
We compute a bootstrap percentile interval over $n$ resamples of the
prompt set and report the $5\text{th}, 50\text{th}, 95\text{th}$ percentiles
as $R_\text{lo}, R_\text{point}, R_\text{hi}$ respectively, i.e.\ a $90\%$
percentile interval centred on the bootstrap median.

\paragraph{Why bootstrap-median rather than plug-in.}
At our operating cell sizes ($n\!\approx\!1500$--$2000$ prompts,
$d\!=\!2048$ Inception features) the plug-in
$R_\text{sel}\!=\!\widehat{W_2^2}(\text{base},\text{mean})/\max\widehat{W_2^2}(\text{base},\{\text{var},\text{scale}\})$
inherits a non-linear functional bias: the Bures functional is
non-linear in the covariance and the bias does not cancel between
numerator and denominator, so the plug-in $R_\text{sel}$ does not
match the bootstrap distribution of the same ratio. Reporting the
bootstrap-median percentiles therefore aligns the headline point and
its interval under a single statistical object; the plug-in value is
retained in our release for audit.
A formal bias decomposition and the audit cells we use to verify the
correction are reported in Appendix~\ref{app:estimator}.

\paragraph{Reading a cell.}
Rather than a binary PASS/NOT-PASS verdict, we report the continuous
ratio $R_\text{point}$ with its $[R_\text{lo},R_\text{hi}]$ interval and
the dominant mode, and compare cells across models under a single
matched estimator. Three distinct uses of the ratio recur and should
not be conflated. (i) $R\!=\!1$ is the \emph{equality line}: the value
at which the mean perturbation exactly matches the strongest
higher-moment perturbation on that layer, used only as a reference
boundary, not a detector. (ii) The \emph{rectified-flow latent
detector} (Section~\ref{sec:exp:rf-detector}) is a \emph{sustained-band}
criterion: a model/configuration passes when its latent-layer ratios have $R_\text{lo}\!>\!1$
on at least three of the four low-to-mid strengths
$s\!\in\!\{0.05,0.1,0.2,0.3\}$, including at least one $s\!\geq\!0.2$;
this is what separates a genuine band from an isolated single-cell
low-$s$ excursion. (iii) The bar $R_\text{lo}\!\geq\!2.0$ with dominant
mode \texttt{mean} is a conservative \emph{strong-selectivity visual
marker} only, not the detector threshold. The analysis in
Section~\ref{sec:experiments} turns on the relative ordering of
CI-disjoint ratios and on the sustained-band criterion, not on the
$R_\text{lo}\!\geq\!2.0$ marker.
All reported intervals quantify prompt-resampling uncertainty under a
fixed checkpoint, sampler, and generation seed; unless explicitly noted
they should not be read as full cross-seed uncertainty intervals, and
they do not
include checkpoint-training or generation-seed variability. Intervals
narrower than $0.005$ round to a single value at two decimal places and
are printed as e.g.\ $0.66\,[0.66,0.66]$.\footnote{A width below $0.005$
collapses to a point under two-decimal rounding; the full-precision
intervals are in the released per-cell tables.}
We do not read isolated cells as discoveries. The main claims are
restricted to structured contrasts stated in advance --- sustained bands
across adjacent strengths, matched-step same-family contrasts, and
CI-separated low-strength peaks --- and we apply no family-wise
multiple-comparison correction; the released grid should be read as
exploratory outside these stated contrasts.

\subsection{Matched Estimator across the Sweep}
\label{sec:method:estimator}

Every ratio reported in this paper comes from a single two-stage
pipeline applied identically to all 23 models. Stage~A computes
per-cell true Bures $W_2^2$ point estimates from an
$n_\text{base}$-prompt conditioning bank against per-strength perturbed
banks of the same size. Stage~B (GPU \texttt{float64}) bootstraps
$n_\text{resample}\!=\!200$ true Bures
$W_2^2$ ratios per cell, replaces the point with the bootstrap median,
and records the $5/95$th percentile interval, so point and CI come from
one coherent distribution. Holding the estimator fixed across SDXL,
SD1.5, SD3.5, PixArt-$\alpha$, and FLUX is what makes the
cross-backbone ratios directly comparable. A faster
diagonal-covariance proxy used for early sweeps is biased relative to
the true Bures estimator and is never used for a main-text claim; the
bias analysis and audit trail are in Appendix~\ref{app:estimator}.

\section{Experiments}
\label{sec:experiments}

\subsection{Experimental Setup}
\label{sec:exp:setup}

\paragraph{The 23-model sweep.}
We run the probe on a sweep of $23$ models, all measured under the
identical matched estimator of Section~\ref{sec:method:estimator}: five
teachers and $18$ distilled students, spanning five backbone families,
three architecture classes (UNet, DiT, MMDiT), and five distillation
paradigms. We count a \emph{model} as a checkpoint with a fixed sampler
step setting, so student step variants (e.g.\ Hyper at $1/4/8$ steps)
count separately; FLUX.1-dev is one checkpoint evaluated at both $28$
and $50$ steps and counted once. Table~\ref{tab:zoo} lists the full set. The design is a
stress test: a layer reading is only an instrument if it tracks the same
factor across families, prediction types, paradigms, and step counts
rather than an artefact of any single configuration.

\begin{table}[t]
\centering
\small
\setlength{\tabcolsep}{4pt}
\begin{tabular}{llllll}
\toprule
\textbf{Family} & \textbf{Arch.\ class} & \textbf{Text enc.} & \textbf{Role} & \textbf{Paradigm} & \textbf{Steps} \\
\midrule
SDXL        & UNet           & 2$\times$CLIP        & teacher & ---                  & 50 \\
SDXL        & UNet           & 2$\times$CLIP        & student & distribution-match.  & 1, 4 \\
SDXL        & UNet           & 2$\times$CLIP        & student & mixed (Hyper)        & 1, 4, 8 \\
SDXL        & UNet           & 2$\times$CLIP        & student & trajectory (LCM)     & 4, 8 \\
SDXL        & UNet           & 2$\times$CLIP        & student & progressive-adv      & 4 \\
SDXL        & UNet           & 2$\times$CLIP        & student & ADD (Turbo)          & 1, 4 \\
\midrule
SD1.5       & UNet           & 1$\times$CLIP        & teacher & ---                  & 50 \\
SD1.5       & UNet           & 1$\times$CLIP        & student & trajectory (LCM)     & 4 \\
SD1.5       & UNet           & 1$\times$CLIP        & student & backward-sim (Flash) & 4 \\
SD1.5       & UNet           & 1$\times$CLIP        & student & mixed (Hyper)        & 4 \\
SD1.5       & UNet           & 1$\times$CLIP        & student & ADD (SD-Turbo)       & 1, 4 \\
\midrule
PixArt-$\alpha$ & DiT, $\epsilon$-pred & T5        & teacher & ---                  & 20 \\
PixArt-$\alpha$ & DiT, $\epsilon$-pred & T5        & student & trajectory (LCM)     & 4 \\
\midrule
SD3.5       & MMDiT, rect.-flow & 2$\times$CLIP + T5 & teacher & ---               & 28 \\
SD3.5       & MMDiT, rect.-flow & 2$\times$CLIP + T5 & student & ADD (Turbo)       & 4 \\
\midrule
FLUX.1-dev  & DiT, rect.-flow & CLIP-L + T5         & teacher & ---                  & 28, 50 \\
FLUX.1      & DiT, rect.-flow & CLIP-L + T5         & student & ADD (schnell)        & 4 \\
\bottomrule
\end{tabular}
\caption{\textbf{The 23-model sweep.} Five teachers and $18$ distilled
students across five backbone families, three architecture classes, and
the five distillation loss families~(a)--(e) of
Section~\ref{sec:related:distill}. The \textbf{Paradigm} column gives the
operational per-checkpoint label; the adversarial family~(a) appears both
as a progressive-adversarial variant (SDXL-Lightning,
\emph{progressive-adv}) and as a single-step variant (\emph{ADD},
Turbo/SD-Turbo), while \emph{trajectory}$=$(b), \emph{distribution-match}
$=$(c), \emph{mixed}$=$(d), and \emph{backward-sim}$=$(e). SDXL
contributes $10$ students spanning families (a)--(d); SD1.5 contributes
$5$ and adds the backward-simulation family (e) via Flash (no public
distribution-matching checkpoint at sweep time), so the two UNet families
together cover all five loss families. The three non-UNet
families contribute one student each, chosen to extend the two strongest
findings off the UNet backbone (see Sections~\ref{sec:exp:rf-detector}
and \ref{sec:exp:spike}). FLUX.1-dev (one model weight) is measured at
both $28$ and $50$ steps as a step-count robustness check and counted
once; the two runs appear as separate rows in the appendix tables.
Here ``teacher'' denotes the high-step reference checkpoint used as the
non-few-step counterpart within a family, not necessarily an undistilled
training teacher in the vendor's release lineage (relevant for
FLUX.1-dev, which is itself guidance-distilled).}
\label{tab:zoo}
\end{table}

\paragraph{Prompts and metrics.}
Unless stated otherwise, measurements start from $2{,}000$ MS-COCO
\citep{lin2014coco} val
captions, fixed across all cells of the perturbation grid (SD1.5 and
SDXL at $512^2$; SD3.5, PixArt-$\alpha$, and FLUX at $1024^2$). After
discarding generations that fail to decode or produce degenerate
features, the effective per-cell bank is $n\!\approx\!1500$--$2000$
prompts (Section~\ref{sec:method:k1}), held fixed within a model. Image
features are Inception-v3 \citep{szegedy2016inception} pool3 ($2048$-d);
the $W_2^2$ distance is the
closed-form Bures functional \citep{bhatia2019bures} under a Gaussian
assumption on Inception
features. Bootstrap CIs use the matched estimator of
Section~\ref{sec:method:estimator} ($n_\text{resample}\!=\!200$,
base-fixed).

\subsection{The Three Layers Read Three Orthogonal Factors}
\label{sec:exp:orthogonal}

\begin{figure}[t]
\centering
\includegraphics[width=\linewidth]{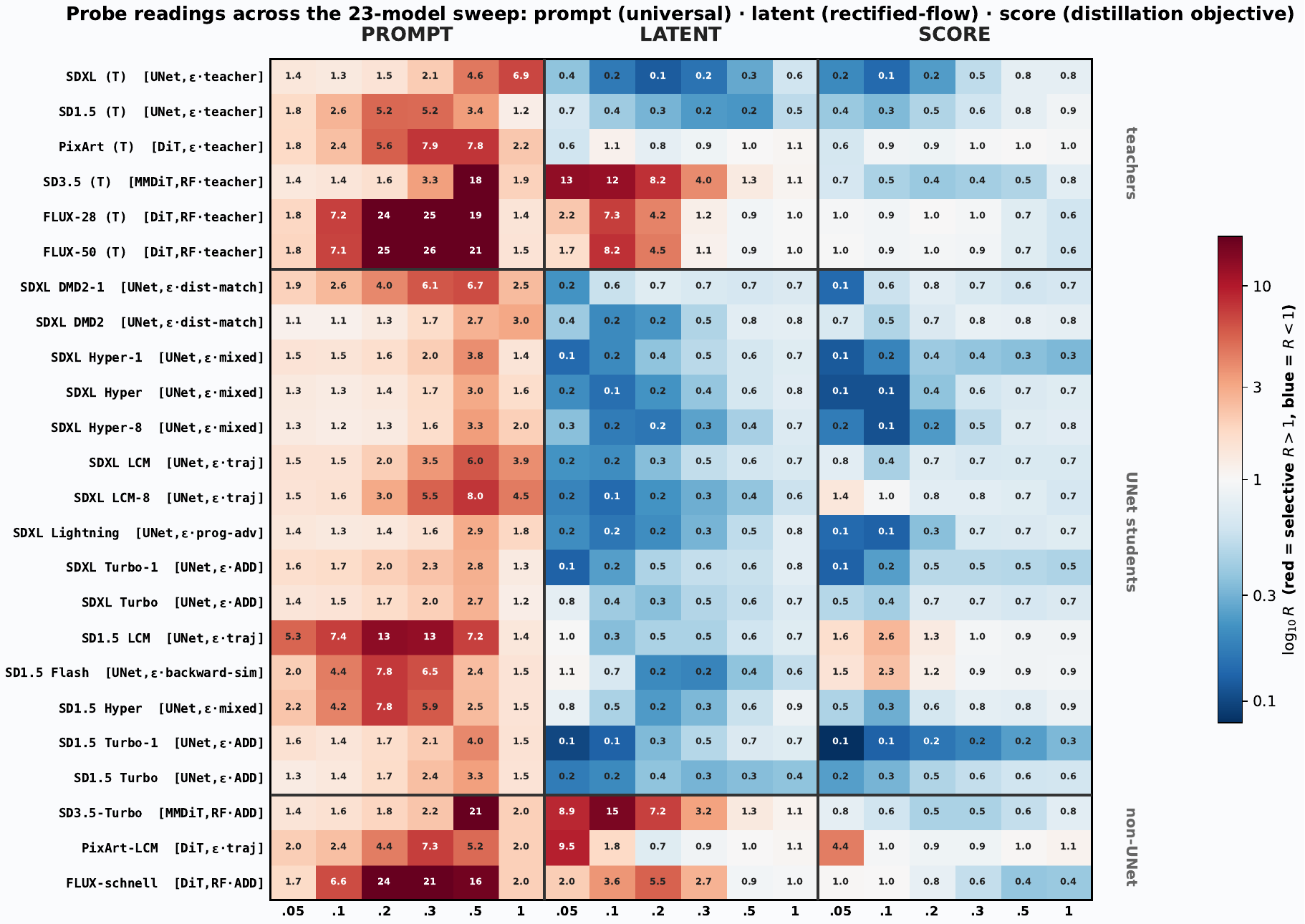}
\caption{\textbf{Master heatmap of the 23-model sweep, all three layers.}
Rows are model/configuration rows (teachers, then UNet students, then
non-UNet students; FLUX.1-dev appears at both $28$ and $50$ steps);
columns are the three layers $\times$ six strengths. Cell colour is
$\log_{10} R$ on a diverging scale centred at $R{=}1$ (red $>1$, blue
$<1$). Three patterns are visible at once: the \textbf{latent} block
turns red only on the rectified-flow rows (SD3.5, FLUX, teacher and
student) --- the P1 fingerprint; the \textbf{score} column at
$s\!=\!0.5$ is lowest (most blue) on the ADD rows --- the P2 detector; and an
isolated red cell at low $s$ on the \textbf{score} layer marks the
trajectory-rollout spike (P3), strongest for PixArt-LCM. The prompt
block is uniformly warm (universal selectivity, discriminates none of our factors of interest).}
\label{fig:master-heatmap}
\end{figure}

The central empirical observation of the paper is that the three
injection layers are not three noisy views of one quantity; each tracks
a distinct, empirically separable property of the model
(Figure~\ref{fig:master-heatmap}). Table~\ref{tab:orthogonal}
summarises the partition that holds across all $23$ models.

\begin{table}[t]
\centering
\small
\setlength{\tabcolsep}{5pt}
\begin{tabular}{lll}
\toprule
\textbf{Layer} & \textbf{Factor it reads} & \textbf{Signature across the 23-model sweep} \\
\midrule
prompt & common prompt-conditioning &
  Prompt-mean selective on \emph{every} model (teacher and \\
 & response (universal here) & student, all five families); peak ratio varies widely \\
 & & ($\sim\!2.7$--$26$) and tracks no factor of interest. \\
\addlinespace
latent & prediction type &
  Sustained low-strength elevation ($R\!>\!1$ across a band) \\
 & (rectified-flow vs $\epsilon$-pred) & only on rectified-flow denoisers; $\epsilon$-pred models \\
 & & lack the band. \\
\addlinespace
score & distillation objective &
  ADD students are the lowest same-family student at the \\
 & (ADD vs non-ADD) & canonical $4$-step comparison; trajectory-rollout students spike at low $s$. \\
\bottomrule
\end{tabular}
\caption{\textbf{The three layers read three empirically separable
factors.} The
prompt layer is a universal prompt-mean response within the tested T2I
diffusion models (a sanity channel, not a discriminator); the latent layer is a near-binary detector of the
prediction type (Section~\ref{sec:exp:rf-detector}); the score layer
reads the distillation objective (Sections~\ref{sec:exp:add-detector}
and \ref{sec:exp:spike}). The probe is therefore a decomposable
instrument, not a single scalar.}
\label{tab:orthogonal}
\end{table}

The prompt-layer reading is universal: every model in the sweep is
prompt-mean selective. Its magnitude varies substantially, however ---
peak prompt-mean ratios span roughly $2.7$ to $26$ across the sweep ---
and it tracks neither the prediction type nor the distillation objective.
This makes the prompt layer
a useful sanity channel but a poor discriminator. The discriminating
power lives in the other two layers, which we examine next. (One
contamination caveat is deferred to Section~\ref{sec:exp:spike}: on
trajectory-rollout students the latent reading is perturbed at the
lowest strengths; the latent factor is therefore read as a
sustained low-to-mid band rather than a single-cell crossing.)

\subsection{Main Finding: the Latent Layer Detects Rectified-Flow, and Survives ADD Distillation}
\label{sec:exp:rf-detector}

The latent layer separates the sweep into a clean, near-binary split
(Figure~\ref{fig:probe-rf}B). Across all
five teachers, only the two rectified-flow backbones (SD3.5, FLUX) show
a \emph{sustained} low-strength latent band with $R$ well above $1$
across several strengths; the three $\epsilon$-pred teachers (SDXL,
SD1.5, PixArt-$\alpha$) do not
(Table~\ref{tab:teacher-latent}). SDXL and SD1.5 stay below $1$ at every
strength; PixArt-$\alpha$ shows only a small near-threshold excursion
(peak $1.13$ at $s\!=\!0.1$), an order of magnitude short of the RF band
and without the sustained low-$s$ shape. Crucially, PixArt-$\alpha$ shares
SD3.5's wide T5 conditioning but reverts the prediction type to
$\epsilon$, and it does \emph{not} reproduce the RF latent band --- which
rules out the hypothesis that wide T5 conditioning alone is sufficient to
produce the band and points to the prediction type / rectified-flow
parameterization as the best-supported factor within this sweep.

\begin{table}[t]
\centering
\small
\setlength{\tabcolsep}{5pt}
\begin{tabular}{llcccc}
\toprule
\textbf{Teacher} & \textbf{Arch.\ / pred.} & \textbf{Cond.} & \textbf{prompt peak} & \textbf{latent @ $s\!=\!0.05$} & \textbf{latent peak} \\
\midrule
SD1.5            & UNet, $\epsilon$        & 1$\times$CLIP        & 5.20 & 0.65 & $<1$ \\
SDXL             & UNet, $\epsilon$        & 2$\times$CLIP        & 6.88 & $<1$ & $<1$ \\
PixArt-$\alpha$  & DiT, $\epsilon$         & T5                  & 7.93 & 0.61 [0.59, 0.62] & 1.13 \\
SD3.5            & MMDiT, \textbf{RF}      & 2$\times$CLIP + T5   & 17.91 & \textbf{13.01 [12.54, 13.43]} & 13.01 \\
FLUX.1-dev (28)  & DiT, \textbf{RF}        & CLIP-L + T5          & 25.44 & \textbf{2.17 [2.07, 2.25]} & 7.31 \\
FLUX.1-dev (50)  & DiT, \textbf{RF}        & CLIP-L + T5          & 26.25 & 1.68 [1.62, 1.75] & 8.21 \\
\bottomrule
\end{tabular}
\caption{\textbf{Latent-layer PASS is a rectified-flow fingerprint.}
Only the two RF teachers show a sustained latent band above $1$ with
large low-$s$ peaks (bold);
the three $\epsilon$-pred teachers do not. SDXL and SD1.5 stay $<1$
throughout; PixArt-$\alpha$ has only a small near-threshold excursion
(peak $1.13$ at $s\!=\!0.1$), far short of the RF band. PixArt-$\alpha$
holds T5 conditioning fixed but is $\epsilon$-pred, falsifying the
wide-conditioning hypothesis. FLUX's 28- and 50-step runs agree, so the
effect is not a step-count artefact. Prompt-peak column shown for
context (magnitude varies and tracks nothing,
Section~\ref{sec:exp:orthogonal}); latent peak occurs at $s\!=\!0.05$ for
SD3.5 and $s\!=\!0.1$ for FLUX.}
\label{tab:teacher-latent}
\end{table}

\paragraph{The fingerprint survives distillation.}
The detector would be a teacher-only curiosity if distillation erased
it. It does not. The two distilled students on RF backbones --- both
trained with a pure adversarial objective (ADD), which (next subsection)
flattens the score-layer signal --- nonetheless retain the latent
fingerprint: SD3.5-Turbo passes with a sustained band ($8.88$ at
$s\!=\!0.05$, peak $15.27$ at $s\!=\!0.1$, still $7.23$ at $s\!=\!0.2$),
and FLUX.1-schnell passes across a band ($2.03$ at $s\!=\!0.05$, peak
$5.53$ at $s\!=\!0.2$). Every
UNet student (SDXL, SD1.5) lacks the band. The one $\epsilon$-pred DiT
student, PixArt-LCM, is the informative boundary case: it has an
isolated latent excursion at the two lowest strengths ($9.53$ at
$s\!=\!0.05$, $1.76$ at $s\!=\!0.1$) that drops below $1$ by $s\!=\!0.2$
and stays near-threshold thereafter, and does not form the sustained RF
band --- this is the same
trajectory-rollout sensitivity that surfaces on its score layer
(Section~\ref{sec:exp:spike}), not the RF signature. Read as a
band-shaped criterion rather than a single low-$s$ crossing, the latent
layer is thus, within this sweep, a \emph{near-binary detector} of
the prediction type that survives ADD distillation, seen across two RF backbone families
(MMDiT and DiT) as both teacher and student --- four RF
cases (Table~\ref{tab:rf-confusion}).

\begin{table}[t]
\centering
\small
\setlength{\tabcolsep}{6pt}
\begin{tabular}{lcc}
\toprule
\textbf{latent ratio (low $s$)} & \textbf{teacher} & \textbf{distilled student} \\
\midrule
rectified-flow      & SD3.5 13.01; FLUX 2.17 \;(band) & SD3.5-Turbo 8.88; schnell 2.03 \;(band) \\
$\epsilon$-pred     & SDXL, SD1.5, PixArt \;(no band)     & all SDXL/SD1.5 students, PixArt-LCM\textsuperscript{$\dagger$} \;(no band) \\
\bottomrule
\end{tabular}
\caption{\textbf{Latent layer as a $2{\times}2$ confusion table: RF vs
$\epsilon$-pred $\times$ teacher vs student.} Classification uses the
sustained low-to-mid latent band, not a single-cell crossing. All four
cells classify correctly; the RF band is present in teachers and
survives ADD distillation.
\textsuperscript{$\dagger$}PixArt-LCM has an isolated latent excursion
at the two lowest strengths ($9.53$ at $s\!=\!0.05$, $1.76$ at
$s\!=\!0.1$) that drops below $1$ by $s\!=\!0.2$ and stays
near-threshold thereafter; it is a
trajectory-rollout artefact, not the sustained RF band
(Section~\ref{sec:exp:spike}).}
\label{tab:rf-confusion}
\end{table}

\subsection{Secondary Finding 1: a Score-Layer ADD-vs-non-ADD Detector}
\label{sec:exp:add-detector}

On the score layer at $s\!=\!0.5$, and \emph{at the canonical $4$-step
comparison}, the
adversarial-distillation (ADD) student occupies the low end of its
family, and is the lowest same-family student at this $4$-step
comparison: the ADD student sits below every $4$-step non-ADD student,
with non-overlapping CIs, on both UNet families that
contain a same-family non-ADD student
(Table~\ref{tab:add-detector}). At $4$ steps on SDXL, Turbo-4 ($0.66$
[$0.66$, $0.66$]) falls below the nearest non-ADD student (LCM $0.68$
[$0.67$, $0.68$]) with disjoint intervals, and below the rest
($0.69$--$0.82$); this SDXL margin is small in absolute magnitude but
CI-disjoint under the matched estimator, whereas on SD1.5 the gap is
wider still (SD-Turbo-4 $0.60$
[$0.60$, $0.60$] vs Hyper $0.77$ [$0.77$, $0.78$]). The
two non-UNet ADD students extend the low-ADD end of the contrast to the
MMDiT and DiT families (SD3.5-Turbo $0.55$, FLUX-schnell $0.38$).

\begin{table}[t]
\centering
\small
\setlength{\tabcolsep}{4pt}
\begin{tabular}{llcc}
\toprule
\textbf{Family} & \textbf{Steps} & \textbf{ADD student, $R(s\!=\!0.5)$ [CI]} & \textbf{nearest non-ADD, same steps [CI]} \\
\midrule
SDXL   & 4 & Turbo-4 $0.66$ [0.66, 0.66] & LCM $0.68$ [0.67, 0.68]; then $0.69$--$0.82$ \\
SD1.5  & 4 & SD-Turbo-4 $0.60$ [0.60, 0.60] & Hyper $0.77$ [0.77, 0.78]; LCM/Flash $0.91$--$0.92$ \\
SD3.5  & 4 & SD3.5-Turbo-4 $0.55$ [0.53, 0.56] & --\textsuperscript{$\ddagger$} \\
FLUX   & 4 & schnell-4 $0.38$ [0.36, 0.39] & --\textsuperscript{$\ddagger$} \\
\bottomrule
\end{tabular}
\caption{\textbf{The score-layer ADD detector, at the canonical
$4$-step comparison.}
Among $4$-step students, the ADD student is the lowest same-family
student on
both UNet families with non-overlapping CIs (the nearest non-ADD
competitor is shown; full per-cell CIs in
Appendix~\ref{app:full-tables}). This is a $4$-step contrast and is
\emph{not} extrapolated to $1$-step, where aggressive non-ADD students
can enter the ADD range. The SDXL margin ($0.66$ vs $0.68$) is
small in absolute magnitude but CI-disjoint under the matched estimator;
the stronger separation is on SD1.5 ($0.60$ vs $0.77$). The two non-UNet ADD
students (SD3.5, FLUX) extend the low-ADD end to MMDiT and DiT.
\textsuperscript{$\ddagger$}SD3.5 and FLUX have no public non-ADD
distilled student, so they contribute ADD points only, not a same-family
contrast.}
\label{tab:add-detector}
\end{table}

\paragraph{Scope: matched-step binary detector, not a five-paradigm
ordering.} Two caveats bound the claim. First, the detector is a
\emph{binary} ADD-vs-rest cut, not a ranking of the five paradigms: the
four non-ADD paradigms are not pairwise separable at $s\!=\!0.5$ (their
$4$-step medians span $0.68$--$0.82$, within $\sim\!0.15$ of one
another), so we do not
order them. Second, the separation is \emph{step-matched}, not absolute:
at $1$ step the most aggressive non-ADD students approach the ADD range
(SDXL Hyper-1 reaches $0.35$, below Turbo-1's $0.49$), because step count
alone moves a student by as much as the paradigm does (Hyper-SDXL spans
$0.35\!\to\!0.72$ across $1/4/8$ steps). The honest reading is therefore
that an adversarial-dominated objective drives the score-layer ratio
down, and at fixed step count this isolates ADD from the rest; the cut
should not be quoted across mismatched step counts.

\subsection{Secondary Finding 2: a Trajectory-Rollout Early-Strength Score Spike}
\label{sec:exp:spike}

\begin{figure}[t]
\centering
\includegraphics[width=0.86\linewidth]{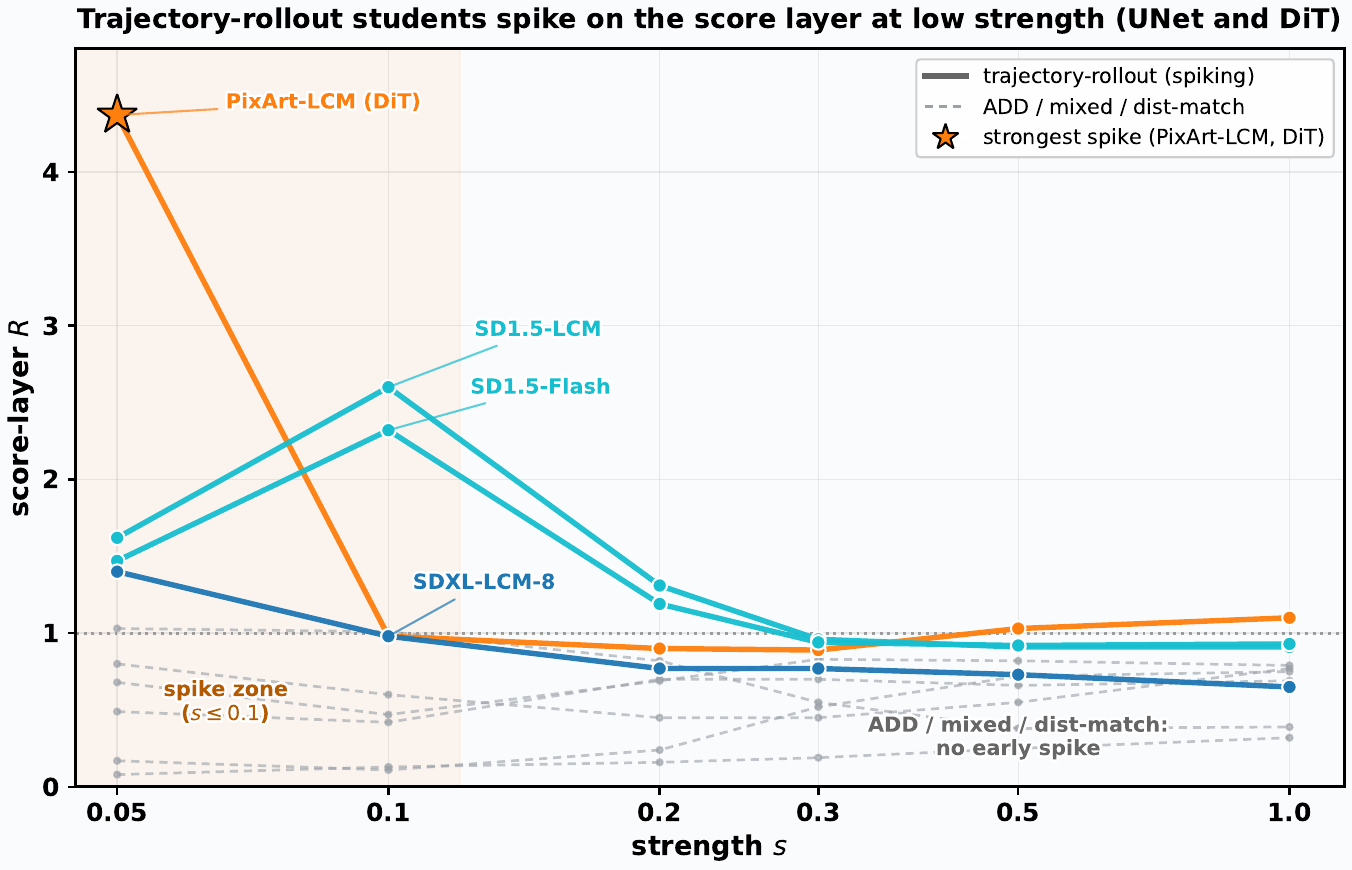}
\caption{\textbf{Trajectory-rollout students spike on the score layer at
low strength.} Score-layer $R$ vs strength $s$ (log $x$). Solid coloured
curves are trajectory-rollout students (LCM on UNet and DiT, Flash on
SD1.5); their score ratio rises above $R{=}1$ at $s\!\leq\!0.1$ and then
falls. Grey dashed curves are the ADD / mixed / distribution-matching
students, none of which have a CI-separated low-$s$ peak above $1$.
PixArt-LCM (DiT) shows the strongest spike
($R\!=\!4.37$ at $s\!=\!0.05$). On DiT/MMDiT, the ADD controls
(SD3.5-Turbo, FLUX-schnell) show no spike, so the spike separates
trajectory-rollout from ADD independently of architecture.}
\label{fig:score-spike}
\end{figure}

The score layer carries a second, shape-based signal. Some students
exhibit a \emph{localized early-strength spike}: the score ratio rises
above $1$ at low strength (with a CI strictly above $1$) and then falls,
so that the low-$s$ peak ($s\!\leq\!0.1$) is CI-separated above the
$s\!=\!0.5$ value (Figure~\ref{fig:score-spike}, Table~\ref{tab:spike}).
Under this criterion the spike appears \emph{only} on students whose loss
trains on student or generated trajectory segments (our operational
\emph{trajectory-rollout} label, which does not imply the absence of
auxiliary adversarial terms): LCM on both UNet backbones (SDXL-LCM at
$8$ steps), Flash (backward-simulation) on SD1.5, and most strongly
PixArt-LCM on a pure DiT backbone (peak $4.37$ [$4.28$, $4.45$] at
$s\!=\!0.05$, the largest distilled spike in the sweep). No tested ADD,
mixed, or distribution-matching representative satisfies the
CI-separated spike criterion: their score ratios have no
CI-separated low-$s$ peak above $1$. Within the full released sweep the
spike is thus exclusive
to trajectory-rollout among the tested paradigms, though not universal within it
(SDXL-LCM at $4$ steps does not reach the threshold, so magnitude is
step- and backbone-dependent). The most tempting counterexample,
FLUX-schnell (ADD), has low-$s$ score $1.03$ [$0.98$, $1.08$] --- a peak
whose CI straddles $1$ --- so it does not meet the criterion; the
DiT/MMDiT ADD controls (SD3.5-Turbo, FLUX-schnell) show no
spike, so on DiT the spike still separates trajectory-rollout from
ADD. The spike is thus associated with trajectory-rollout objectives and
observed on both UNet and DiT, while its appearance and magnitude shift
with step count and backbone.

\begin{table}[t]
\centering
\small
\setlength{\tabcolsep}{6pt}
\begin{tabular}{llccc}
\toprule
\textbf{Backbone $\times$ paradigm} & \textbf{Arch.} & \textbf{low-$s$ peak [CI]} & $s\!=\!0.50$ \textbf{[CI]} & \textbf{spike?} \\
\midrule
PixArt $\times$ LCM-4        & DiT  & \textbf{4.37} [4.28, 4.45] & 1.03 [1.03, 1.03] & \checkmark\checkmark \\
SD1.5 $\times$ LCM-4         & UNet & \textbf{2.60} [2.59, 2.61] & 0.91 [0.91, 0.92] & \checkmark \\
SD1.5 $\times$ Flash-4       & UNet & \textbf{2.32} [2.30, 2.34] & 0.92 [0.92, 0.92] & \checkmark \\
SDXL $\times$ LCM-8          & UNet & \textbf{1.40} [1.35, 1.45] & 0.73 [0.73, 0.73] & \checkmark \\
FLUX $\times$ schnell-4 (ADD) & DiT  & 1.03 [0.98, 1.08] & 0.38 [0.36, 0.39] & $\times$\textsuperscript{$\ast$} \\
SDXL $\times$ LCM-4          & UNet & 0.76 [0.74, 0.78] & 0.68 [0.67, 0.68] & $\times$ \\
SDXL $\times$ Lightning-4    & UNet & 0.14 & 0.69 & $\times$ \\
SD1.5 $\times$ Hyper-4       & UNet & 0.46 & 0.77 & $\times$ \\
SDXL $\times$ Hyper-8        & UNet & 0.17 & 0.72 & $\times$ \\
SDXL $\times$ DMD2-1         & UNet & 0.58 & 0.62 & $\times$ \\
SDXL $\times$ Turbo-4 (ADD)  & UNet & 0.49 & 0.66 & $\times$ \\
SD1.5 $\times$ SD-Turbo-4 (ADD) & UNet & 0.31 & 0.60 & $\times$ \\
SD3.5 $\times$ Turbo-4 (ADD) & MMDiT & 0.80 & 0.55 & $\times$ \\
\bottomrule
\end{tabular}
\caption{\textbf{Score-layer early-strength spike, by paradigm.} Spike
criterion: the low-$s$ peak ($s\!\leq\!0.1$) has a CI strictly above $1$
\emph{and} is CI-separated above the $s\!=\!0.5$ value (CIs shown for the
four positive rows and the FLUX-schnell counterexample; the negative rows
fall well below $1$ at low $s$). Under this rule
the spike appears \emph{only} in trajectory-rollout paradigms, across
both UNet and DiT. PixArt-LCM (DiT) shows the strongest spike.
\textsuperscript{$\ast$}FLUX-schnell (ADD) has a low-$s$ score of $1.03$
[$0.98$, $1.08$] whose CI straddles $1$, so it does not meet the
criterion --- the naive rule $R(s\!\leq\!0.1)>R(s\!=\!0.5)$ alone would
misclassify it, which is why the CI-separated form is used. The $13$
rows are one representative configuration per family\,$\times$\,paradigm
(at a representative step count), plus a second SDXL-LCM step variant
($4$ vs $8$) to show step-dependence; the remaining step-variants of the
$18$-student sweep, all non-spiking, and full per-cell values with CIs
are in the released tables.}
\label{tab:spike}
\end{table}

\paragraph{Latent contamination caveat.}
PixArt-LCM also shows a latent excursion at the two lowest strengths
($9.53$ at $s\!=\!0.05$, $1.76$ at $s\!=\!0.1$) co-located with its score
spike. This is \emph{not} the band-shaped RF latent signature of
Section~\ref{sec:exp:rf-detector} (which is sustained across
$s\!\in\![0.05, 0.3]$ on SD3.5/FLUX); it is the same trajectory-rollout
sensitivity surfacing in both layers at low strength, and it collapses
below $1$ by $s\!=\!0.2$. Because the RF signature is read as a
\emph{sustained band} rather than a single low-$s$ crossing, this
isolated two-cell excursion does not classify PixArt-LCM as
rectified-flow (Table~\ref{tab:rf-confusion}).

\subsection{Distillation Amplifies Prompt Collapse Strongly on SD1.5, Weakly Elsewhere}
\label{sec:exp:amplification}

A final observation uses the architecture-matched teacher/student pairs.
Comparing the prompt-layer peak ratio of each student against its own
teacher, distillation amplifies prompt-mean collapse strongly on SD1.5,
only weakly and inconsistently on SDXL, and near-neutrally on the
DiT/MMDiT pairs (Table~\ref{tab:amplification}). We report this as an
observation rather than a law: the comparison is peak-versus-peak (the
peaks sit at different strengths on teacher and student), SDXL's
students straddle their teacher (the amplification is carried by SD1.5),
and the non-UNet families contribute a single student each.

\begin{table}[t]
\centering
\small
\setlength{\tabcolsep}{6pt}
\begin{tabular}{llc}
\toprule
\textbf{Family} & \textbf{Arch.} & \textbf{best student peak / teacher peak} \\
\midrule
SD1.5           & UNet            & 13.32 / 5.20 = \textbf{2.56$\times$} \\
SDXL            & UNet            & 7.98 / 6.88 = 1.16$\times$\textsuperscript{$\S$} \\
SD3.5           & MMDiT           & 21.05 / 17.91 = 1.18$\times$ \\
PixArt-$\alpha$ & DiT             & 7.30 / 7.93 = 0.92$\times$ \\
FLUX            & DiT             & 23.66 / 26.25 = 0.90$\times$ \\
\bottomrule
\end{tabular}
\caption{\textbf{Prompt-collapse amplification by distillation.} The
best-student/teacher prompt-peak ratio is large only on SD1.5
($2.56\times$); SDXL is near-neutral ($1.16\times$ for its strongest
student, with most SDXL students at or below their teacher) and DiT/MMDiT
are near or below $1$. Observation only (peak-vs-peak; SDXL carries $10$
students, the other four families one student each).
\textsuperscript{$\S$}SDXL best student is LCM-8; the SDXL student
prompt peaks span $\sim\!2.7$--$8.0$, straddling the teacher's $6.88$.}
\label{tab:amplification}
\end{table}

\section{Discussion}
\label{sec:discussion}

\subsection{What the probe establishes, and what it does not}
\label{sec:discussion:claims}

The three layers read three empirically separable factors
(Section~\ref{sec:exp:orthogonal}), and the strongest of the three
readings is the latent layer. The rectified-flow detector
(Section~\ref{sec:exp:rf-detector}) rests on a clean separation (two RF
backbones pass, three $\epsilon$-prediction backbones do not), a control
that removes the obvious confound (PixArt-$\alpha$ holds T5 conditioning
fixed but is $\epsilon$-prediction and does not pass), and four
RF cases spanning teacher/student and MMDiT/DiT. We state
this as a property of the \emph{prediction type} that survives the
distillation runs we tested --- both pure-adversarial (ADD) --- and not
as a claim about every possible RF distillation: we did not test a
non-ADD RF student, so ``survives ADD'' is the precise statement. We do
not claim that this criterion is a universal rectified-flow detector
outside the tested public checkpoints; it is the factor that cleanly
separates this sweep.

The score-layer findings are deliberately scoped narrower. Finding~2 is
a \emph{binary} ADD-vs-non-ADD detector, not a five-paradigm ranking:
the four non-ADD paradigms are not pairwise separable at $s\!=\!0.5$ and
step count confounds any finer ordering (Section~\ref{sec:exp:add-detector}).
Finding~3 is associated with trajectory-rollout objectives and observed
on both UNet and DiT, but its appearance and magnitude depend on step
count and backbone (SDXL-LCM spikes at $8$ steps but not $4$), and each
non-UNet paradigm is represented by a
single student. The amplification observation
(Section~\ref{sec:exp:amplification}) is a peak-versus-peak comparison on
single non-UNet students and is reported as an observation, not a law.

\subsection{Limitations}
\label{sec:discussion:limits}

The probe is white-box: it requires the model weights and forward code
to install the hooks, so it does not perform image-only detection. The
non-UNet coverage is three students (ADD on MMDiT and DiT; trajectory on
DiT); we have no non-ADD RF student and no distribution-matching or
mixed student off UNet, so the latent fingerprint is shown to survive
ADD only, and the score-layer paradigm structure is densest on UNet.
Most students are measured at a single seed (cross-seed robustness was
established on the SDXL zoo in earlier work). One reading is
contaminated at low strength: trajectory-rollout students show a
spurious latent excursion at the two lowest strengths, which is why the
RF fingerprint is read as a sustained low-to-mid band rather than a
single low-$s$ crossing (Section~\ref{sec:exp:spike}).

\subsection{Toward a downstream use}
\label{sec:discussion:sowhat}

This paper stops at the instrument and its readings. Two downstream uses
are natural but deliberately left to future work. First, the score layer
reads a property --- the distillation objective --- that is \emph{not}
recoverable from the weights or configuration, so auditing an
undisclosed training recipe from a released checkpoint is a non-circular
application; by contrast, detecting the architecture via the latent
layer is circular, since the prediction type is already declared in the
model configuration, and we do not claim it as an application. Second,
the per-layer readings could serve as a during-training signal to guide
distillation. Both require an evaluation we do not undertake here, and
we make no downstream claim in this version.

\section{Conclusion}
\label{sec:conclusion}

We presented a decomposable, layer-/mode-resolved probe for
diffusion text-to-image models, and applied it under a single matched
estimator to a $23$-model sweep spanning five backbone families, three
architecture classes, and five distillation paradigms. The probe's three
injection layers turn out to read three empirically separable factors --- a
universal prompt-conditioning response on the prompt layer, the prediction
type on the latent layer, and the distillation objective on the score
layer. Within this sweep the latent layer in particular is a
near-binary detector of rectified-flow backbones that survives ADD
distillation, under the
sustained-band criterion, with the obvious wide-T5-conditioning
explanation ruled out by a matched $\epsilon$-prediction control and four
RF cases across two architecture families and teacher/student roles; two further cross-family findings --- a binary
ADD-vs-non-ADD score detector and a trajectory-rollout early-strength
spike present on both UNet and DiT --- complete the picture. We release
all per-cell CI-citable tables and the matched estimator. Turning these
readings into a downstream use --- auditing an undisclosed distillation
recipe, or guiding distillation during training --- is the natural next
step.

\bibliographystyle{plainnat}
\bibliography{refs}

\appendix

\section{Bures Estimator: Plug-in Bias and Bootstrap-Median Fix}
\label{app:estimator}

This appendix records the estimator decision behind every $R_\text{lo}$
number reported in the main text. The squared-Wasserstein-2 ratio
$R_\text{sel}$ defined in eq.~\ref{eq:ratio} is evaluated on the true
Bures $W_2^2$ functional applied to Inception-v3 pool3 statistics,
computed via a low-rank Gram form, on every one of the 23 models in
the sweep. At our operating cell sizes ($n\!\approx\!1500$--$2000$,
$d\!=\!2048$) the plug-in estimator
$W_2^2(\widehat{\mu},\widehat{\Sigma})$ inherits a non-linear
functional bias: because the Bures functional is non-linear in the
covariance, its plug-in does not coincide with the bootstrap
distribution of the same functional, so the plug-in point estimate
does not in general fall inside the bootstrap percentile interval; we
therefore adopt
the bootstrap-median percentile estimator below as the
single-distribution headline.

\paragraph{Bootstrap-median estimator.}
We replace the plug-in with a bootstrap-median percentile estimator:
$n_\text{resample}$ resamples of the
prompt set, the median of the resampled ratios is reported as
$R_\text{point}$, and the $5/95$th percentiles as
$[R_\text{lo}, R_\text{hi}]$, i.e.\ a $90\%$ percentile interval. This
is the headline statistic on every model in the main text. The plug-in
value is retained in our release for
audit purposes; it is never the bar a claim is checked against. All
cross-backbone comparison cells use the base-fixed bootstrap-median
estimator with $n_\text{resample}\!=\!200$ and a fixed seed, applied
identically to every model in the sweep so that all ratios are
estimator-matched.

\paragraph{Matched across backbones.}
The single methodological requirement for the cross-backbone claims is
that the \emph{same} estimator is applied to every model. We confirm
that SDXL, SD1.5, SD3.5, PixArt-$\alpha$, and FLUX teachers and all 18
students are computed under base-fixed bootstrap-median true Bures
$W_2^2$. For the record, a fast diagonal-covariance proxy was used for
early exploratory sweeps but for \emph{no} cell in this paper: it drops
the cross-dimension covariance terms and biases the selectivity ratio
relative to the true Bures functional, with a direction that depends on
the cell (the sign is not uniform across layers/strengths). Rather than
rely on any fixed-sign argument, every cross-backbone cell is recomputed
with the true functional so that all $23$ models are estimator-comparable.

\section{Architecture / Paradigm Master Mapping}
\label{app:taxonomy}

Table~\ref{tab:app-2x2} records the $2\!\times\!2$ disambiguation that
attributes the latent fingerprint to the prediction type rather than the
text-conditioning width: PixArt-$\alpha$ is the
$(\epsilon\text{-pred}, \text{T5})$ cell that falsifies the
wide-conditioning hypothesis.

\begin{table}[h]
\centering
\small
\setlength{\tabcolsep}{6pt}
\begin{tabular}{lcc}
\toprule
 & \textbf{$\epsilon$-prediction} & \textbf{rectified-flow} \\
\midrule
\textbf{T5 (+ CLIP)} & PixArt-$\alpha$: no band (peak $1.13$) ($\times$) & SD3.5: band, peak $13.01$ (\checkmark); FLUX: band, peak $7.31$ ($28$-step) (\checkmark) \\
\textbf{CLIP only}   & SDXL, SD1.5: no band ($\times$)        & (none in sweep) \\
\bottomrule
\end{tabular}
\caption{\textbf{Prediction-type $\times$ conditioning $2{\times}2$.}
Latent-layer PASS is defined by a sustained low-to-mid latent band, not
by a single low-$s$ threshold crossing; it tracks the rectified-flow
column, not the T5 row: PixArt-$\alpha$ has T5 but $\epsilon$-prediction
and does not form the band, ruling out wide conditioning as the cause.}
\label{tab:app-2x2}
\end{table}

\section{Per-Backbone Full Sweep Tables}
\label{app:full-tables}

All cells are $R_\text{point}$ with $[R_\text{lo}, R_\text{hi}]$ in
brackets, base-fixed bootstrap-median true Bures $W_2^2$,
$n_\text{resample}\!=\!200$, full $3$-layer $\times$ $6$-strength grid.
Latent and score columns give $R_\text{point}$ (CIs in the released
JSON). We group by family; teachers first.

\paragraph{Non-UNet teachers: RF backbones (SD3.5, FLUX) and the
$\epsilon$-prediction control (PixArt-$\alpha$).}
Table~\ref{tab:app-rf-teachers} gives the non-UNet teachers: SD3.5 and
FLUX as the rectified-flow cases, and PixArt-$\alpha$ as the matched T5
$\epsilon$-prediction control. The latent column is the
load-bearing one for Section~\ref{sec:exp:rf-detector}.

\begin{table}[h]
\centering
\small
\setlength{\tabcolsep}{4pt}
\begin{tabular}{llcccccc}
\toprule
\textbf{Teacher} & \textbf{layer} & $s\!=\!0.05$ & $s\!=\!0.1$ & $s\!=\!0.2$ & $s\!=\!0.3$ & $s\!=\!0.5$ & $s\!=\!1.0$ \\
\midrule
\multirow{3}{*}{SD3.5 (28, RF)}
 & prompt & 1.36 & 1.42 & 1.59 & 3.27 & \textbf{17.91} & 1.93 \\
 & latent & \textbf{13.01} & 12.37 & 8.22 & 3.96 & 1.30 & 1.08 \\
 & score  & 0.66 & 0.45 & 0.39 & 0.42 & 0.47 & 0.77 \\
\midrule
\multirow{3}{*}{PixArt-$\alpha$ (20, $\epsilon$)}
 & prompt & 1.76 & 2.41 & 5.64 & \textbf{7.93} & 7.85 & 2.18 \\
 & latent & 0.61 & 1.13 & 0.80 & 0.90 & 1.00 & 1.05 \\
 & score  & 0.64 & 0.91 & 0.89 & 0.98 & 1.02 & 0.97 \\
\midrule
\multirow{3}{*}{FLUX.1-dev (28, RF)}
 & prompt & 1.81 & 7.17 & 24.14 & \textbf{25.44} & 19.38 & 1.45 \\
 & latent & 2.17 & 7.31 & 4.23 & 1.18 & 0.93 & 0.99 \\
 & score  & 0.98 & 0.93 & 1.00 & 0.96 & 0.73 & 0.61 \\
\midrule
\multirow{3}{*}{FLUX.1-dev (50, RF)}
 & prompt & 1.84 & 7.12 & 24.95 & \textbf{26.25} & 20.92 & 1.49 \\
 & latent & 1.68 & 8.21 & 4.46 & 1.12 & 0.93 & 0.99 \\
 & score  & 0.99 & 0.93 & 0.96 & 0.93 & 0.74 & 0.60 \\
\bottomrule
\end{tabular}
\caption{\textbf{Rectified-flow and $\epsilon$-control teachers.} A
sustained latent band $>1$ appears only for the two RF teachers;
PixArt-$\alpha$ (matched T5 conditioning, $\epsilon$-prediction) has only
a small near-threshold excursion (peak $1.13$ at $s\!=\!0.1$) and does
not reproduce the band. FLUX 28- and 50-step agree. Entries are
$R_\text{point}$; the lower CI bounds $R_\text{lo}$ at the detector
strengths $s\!\in\!\{0.05,0.1,0.2,0.3\}$ used by the sustained-band
criterion are in the released per-cell tables.}
\label{tab:app-rf-teachers}
\end{table}

\paragraph{Non-UNet students (SD3.5-Turbo, PixArt-LCM, FLUX-schnell).}
Table~\ref{tab:app-nonunet-students} gives the three non-UNet distilled
students. The two ADD students (SD3.5-Turbo, FLUX-schnell) retain the
latent fingerprint band; PixArt-LCM shows the low-strength latent
excursion ($9.53$ at $s\!=\!0.05$, $1.76$ at $s\!=\!0.1$) flagged in
Section~\ref{sec:exp:spike} and the strongest
score early-spike in the sweep.

\begin{table}[h]
\centering
\small
\setlength{\tabcolsep}{4pt}
\begin{tabular}{llcccccc}
\toprule
\textbf{Student} & \textbf{layer} & $s\!=\!0.05$ & $s\!=\!0.1$ & $s\!=\!0.2$ & $s\!=\!0.3$ & $s\!=\!0.5$ & $s\!=\!1.0$ \\
\midrule
\multirow{3}{*}{SD3.5-Turbo-4 (ADD)}
 & prompt & 1.45 & 1.57 & 1.84 & 2.21 & \textbf{21.05} & 1.99 \\
 & latent & 8.88 & \textbf{15.27} & 7.23 & 3.21 & 1.32 & 1.11 \\
 & score  & 0.80 & 0.60 & 0.45 & 0.45 & 0.55 & 0.77 \\
\midrule
\multirow{3}{*}{PixArt-LCM-4 (traj.)}
 & prompt & 2.04 & 2.36 & 4.38 & \textbf{7.30} & 5.15 & 2.00 \\
 & latent & 9.53 & 1.76 & 0.74 & 0.91 & 1.01 & 1.05 \\
 & score  & \textbf{4.37} & 0.98 & 0.90 & 0.89 & 1.03 & 1.10 \\
\midrule
\multirow{3}{*}{FLUX-schnell-4 (ADD)}
 & prompt & 1.74 & 6.56 & \textbf{23.66} & 21.05 & 16.15 & 1.98 \\
 & latent & 2.03 & 3.61 & \textbf{5.53} & 2.73 & 0.93 & 0.97 \\
 & score  & 1.03 & 1.01 & 0.82 & 0.55 & 0.38 & 0.39 \\
\bottomrule
\end{tabular}
\caption{\textbf{Non-UNet distilled students.} SD3.5-Turbo and
FLUX-schnell (both ADD) keep a sustained latent band $>1$ at low-to-mid
$s$; PixArt-LCM's latent excursion ($9.53$ at $s\!=\!0.05$, $1.76$ at
$s\!=\!0.1$) is the rollout artefact and drops below $1$ by
$s\!=\!0.2$ (near-threshold thereafter), so it does not form the
sustained RF band. Score row: ADD
students low with no spike, PixArt-LCM spikes at $s\!=\!0.05$.}
\label{tab:app-nonunet-students}
\end{table}

\paragraph{UNet teachers and students (SDXL, SD1.5).}
Tables~\ref{tab:app-sdxl} and \ref{tab:app-sd15} give the full per-cell
$R_\text{point}$ grid for the two UNet families. Every latent row stays
below $1$ at low strength (no rectified-flow fingerprint,
Section~\ref{sec:exp:rf-detector}); the score column at $s\!=\!0.5$
carries the ADD-detector ordering (Section~\ref{sec:exp:add-detector}),
and the SDXL-LCM-8 / SD1.5-LCM / SD1.5-Flash score rows carry the
low-strength spike (Section~\ref{sec:exp:spike}).

\begin{table}[h]
\centering
\small
\setlength{\tabcolsep}{4pt}
\begin{tabular}{llcccccc}
\toprule
\textbf{Model (SDXL)} & \textbf{layer} & $s\!=\!0.05$ & $s\!=\!0.1$ & $s\!=\!0.2$ & $s\!=\!0.3$ & $s\!=\!0.5$ & $s\!=\!1.0$ \\
\midrule
\multirow{3}{*}{SDXL teacher (50)}
 & prompt & 1.37 & 1.33 & 1.48 & 2.08 & 4.58 & 6.88 \\
 & latent & 0.36 & 0.17 & 0.12 & 0.16 & 0.27 & 0.61 \\
 & score  & 0.20 & 0.14 & 0.25 & 0.52 & 0.80 & 0.80 \\
\midrule
\multirow{3}{*}{SDXL DMD2-1}
 & prompt & 1.94 & 2.56 & 4.04 & 6.08 & 6.65 & 2.51 \\
 & latent & 0.22 & 0.57 & 0.73 & 0.68 & 0.66 & 0.69 \\
 & score  & 0.14 & 0.58 & 0.77 & 0.68 & 0.62 & 0.65 \\
\midrule
\multirow{3}{*}{SDXL DMD2-4}
 & prompt & 1.12 & 1.13 & 1.31 & 1.71 & 2.74 & 3.03 \\
 & latent & 0.40 & 0.20 & 0.23 & 0.47 & 0.76 & 0.78 \\
 & score  & 0.68 & 0.47 & 0.69 & 0.83 & 0.82 & 0.79 \\
\midrule
\multirow{3}{*}{SDXL Hyper-1}
 & prompt & 1.49 & 1.46 & 1.63 & 2.02 & 3.84 & 1.45 \\
 & latent & 0.14 & 0.20 & 0.36 & 0.51 & 0.64 & 0.74 \\
 & score  & 0.12 & 0.19 & 0.40 & 0.37 & 0.35 & 0.32 \\
\midrule
\multirow{3}{*}{SDXL Hyper-4}
 & prompt & 1.32 & 1.26 & 1.36 & 1.71 & 2.99 & 1.65 \\
 & latent & 0.21 & 0.15 & 0.22 & 0.38 & 0.64 & 0.75 \\
 & score  & 0.11 & 0.11 & 0.36 & 0.63 & 0.69 & 0.73 \\
\midrule
\multirow{3}{*}{SDXL Hyper-8}
 & prompt & 1.29 & 1.23 & 1.30 & 1.65 & 3.33 & 1.97 \\
 & latent & 0.35 & 0.17 & 0.16 & 0.26 & 0.43 & 0.71 \\
 & score  & 0.17 & 0.11 & 0.24 & 0.52 & 0.72 & 0.75 \\
\midrule
\multirow{3}{*}{SDXL LCM-4}
 & prompt & 1.51 & 1.51 & 2.03 & 3.48 & 6.01 & 3.85 \\
 & latent & 0.22 & 0.21 & 0.34 & 0.54 & 0.63 & 0.68 \\
 & score  & 0.76 & 0.44 & 0.72 & 0.67 & 0.68 & 0.68 \\
\midrule
\multirow{3}{*}{SDXL LCM-8}
 & prompt & 1.49 & 1.57 & 3.00 & 5.51 & 7.98 & 4.45 \\
 & latent & 0.19 & 0.14 & 0.22 & 0.29 & 0.37 & 0.58 \\
 & score  & \textbf{1.40} & 0.98 & 0.77 & 0.77 & 0.73 & 0.65 \\
\midrule
\multirow{3}{*}{SDXL Lightning-4}
 & prompt & 1.38 & 1.32 & 1.35 & 1.56 & 2.91 & 1.82 \\
 & latent & 0.21 & 0.16 & 0.21 & 0.31 & 0.53 & 0.76 \\
 & score  & 0.14 & 0.13 & 0.35 & 0.68 & 0.69 & 0.73 \\
\midrule
\multirow{3}{*}{SDXL Turbo-1}
 & prompt & 1.61 & 1.69 & 1.97 & 2.29 & 2.81 & 1.31 \\
 & latent & 0.13 & 0.25 & 0.46 & 0.55 & 0.57 & 0.77 \\
 & score  & 0.13 & 0.25 & 0.50 & 0.50 & 0.49 & 0.46 \\
\midrule
\multirow{3}{*}{SDXL Turbo-4}
 & prompt & 1.41 & 1.49 & 1.70 & 1.99 & 2.74 & 1.21 \\
 & latent & 0.79 & 0.41 & 0.35 & 0.48 & 0.55 & 0.71 \\
 & score  & 0.49 & 0.42 & 0.70 & 0.70 & \textbf{0.66} & 0.69 \\
\bottomrule
\end{tabular}
\caption{\textbf{SDXL full sweep.} Teacher $+$ $10$ students, all five
paradigms. No sustained RF-like latent band (all $\epsilon$-pred: latent
$<1$ at low $s$ throughout). Score $s\!=\!0.5$: Turbo-4 ($0.66$, bold) is
the lowest $4$-step SDXL student. SDXL-LCM-8 score at $s\!=\!0.05$
($1.40$, bold) is the SDXL trajectory spike.}
\label{tab:app-sdxl}
\end{table}

\begin{table}[t]
\centering
\small
\setlength{\tabcolsep}{4pt}
\begin{tabular}{llcccccc}
\toprule
\textbf{Model (SD1.5)} & \textbf{layer} & $s\!=\!0.05$ & $s\!=\!0.1$ & $s\!=\!0.2$ & $s\!=\!0.3$ & $s\!=\!0.5$ & $s\!=\!1.0$ \\
\midrule
\multirow{3}{*}{SD1.5 teacher (50)}
 & prompt & 1.75 & 2.65 & 5.20 & 5.16 & 3.37 & 1.20 \\
 & latent & 0.65 & 0.41 & 0.31 & 0.24 & 0.23 & 0.52 \\
 & score  & 0.39 & 0.33 & 0.47 & 0.61 & 0.80 & 0.91 \\
\midrule
\multirow{3}{*}{SD1.5 LCM-4}
 & prompt & 5.34 & 7.38 & 13.32 & 12.80 & 7.18 & 1.38 \\
 & latent & 0.99 & 0.34 & 0.45 & 0.45 & 0.61 & 0.74 \\
 & score  & 1.62 & \textbf{2.60} & 1.31 & 0.96 & 0.91 & 0.91 \\
\midrule
\multirow{3}{*}{SD1.5 Flash-4}
 & prompt & 2.02 & 4.38 & 7.79 & 6.48 & 2.42 & 1.50 \\
 & latent & 1.05 & 0.67 & 0.20 & 0.19 & 0.36 & 0.55 \\
 & score  & 1.47 & \textbf{2.32} & 1.19 & 0.94 & 0.92 & 0.93 \\
\midrule
\multirow{3}{*}{SD1.5 Hyper-4}
 & prompt & 2.25 & 4.24 & 7.77 & 5.89 & 2.50 & 1.49 \\
 & latent & 0.83 & 0.45 & 0.24 & 0.32 & 0.57 & 0.87 \\
 & score  & 0.46 & 0.29 & 0.64 & 0.78 & 0.77 & 0.86 \\
\midrule
\multirow{3}{*}{SD1.5 SD-Turbo-1}
 & prompt & 1.55 & 1.41 & 1.66 & 2.06 & 4.03 & 1.51 \\
 & latent & 0.10 & 0.13 & 0.33 & 0.49 & 0.69 & 0.73 \\
 & score  & 0.08 & 0.13 & 0.16 & 0.19 & 0.25 & 0.32 \\
\midrule
\multirow{3}{*}{SD1.5 SD-Turbo-4}
 & prompt & 1.27 & 1.35 & 1.71 & 2.40 & 3.29 & 1.47 \\
 & latent & 0.17 & 0.20 & 0.36 & 0.31 & 0.34 & 0.36 \\
 & score  & 0.25 & 0.31 & 0.47 & 0.55 & \textbf{0.60} & 0.63 \\
\bottomrule
\end{tabular}
\caption{\textbf{SD1.5 full sweep.} Teacher $+$ $5$ students. No
sustained RF-like latent band; occasional near-threshold low-$s$ latent
values (e.g.\ Flash $1.05$ at $s\!=\!0.05$) do not reproduce the RF
signature. Score $s\!=\!0.5$: SD-Turbo-4 ($0.60$, bold)
is the lowest $4$-step SD1.5 student (ADD detector). SD1.5-LCM and
SD1.5-Flash score peaks at $s\!=\!0.1$ ($2.60$, $2.32$, bold) are the
UNet trajectory spikes.}
\label{tab:app-sd15}
\end{table}

\end{document}